\documentclass{article}

\usepackage[english]{babel}
\usepackage{float}
\usepackage[letterpaper,top=2cm,bottom=2cm,left=3cm,right=3cm,marginparwidth=1.75cm]{geometry}
\usepackage{amsmath}
\usepackage{wrapfig}
\usepackage{array}
\usepackage{graphicx}
\usepackage[colorlinks=true, allcolors=blue]{hyperref}
\usepackage{caption}
\usepackage{subcaption}

\title{\textbf{Efficient Training in Multi-Agent Reinforcement Learning: A Communication-Free Framework for the Box-Pushing Problem}}
\author{
    \begin{minipage}[t]{0.45\textwidth}
        \centering
        David Ge\\
        Department of Computer Science\\
        University of California, Berkeley\\
        2301 Durant Ave, Berkeley, CA 94704\\
        \texttt{lwg0320@berkeley.edu}
    \end{minipage}
    \hspace{0.05\textwidth}
    \begin{minipage}[t]{0.45\textwidth}
        \centering
        Hao Ji\thanks{Corresponding author}\\
        Center for Advanced Research Computing\\
        University of Southern California\\
        3434 South Grand Avenue, Building CAL, Los Angeles, CA 90089\\
        \texttt{haoji@usc.edu}
    \end{minipage}
}

\begin{document}

\date{}
\maketitle

\begin{abstract}
Self-organizing systems consists of autonomous agents that can perform complex tasks and adapt to dynamic environments without a central controller. Prior research often relies on reinforcement learning to enable agents to gain the skills needed for task completion, such as in the box-pushing environment. However, when agents push from opposing directions during exploration, they tend to exert equal and opposite forces on the box, resulting in minimal displacement and inefficient training. This paper proposes a model called Shared Pool of Information (SPI), which enables information to be accessible to all agents and faciliate coordinate and reduce force conflicts among agents, thus enhancing exploration efficiency. Through computer simulations, we demonstrate that SPI not only expedites the training process but also requires fewer steps per episode, significantly improving the agents' collaborative effectiveness.
\end{abstract}

\section{Introduction}

Self-organizing systems can address scalability, bottlenecks, and reliability issues that are common in systems with a central controller \cite{Gorodetskii}. However, in practice, agents that learn purely by themselves without the help of a central entity can result in the agent having unbalanced data \cite{kairouz}. Agents can combat the issue through implicit observations or through direct communication (\cite{claus1998dynamics}, \cite{8039194}, \cite{Oroojlooy2023}). To add the ability to communicate would require additional overhead to the model. The solution proposed in this paper requires no communication or significant overhead.

Cooperation and coordination in reinforcement learning has been a difficult yet crucial problem to solve. There are many studies that aim to find an efficient solution that will allow agents in a multi-agent environment to coordinate. For example, Yexin et al. proposed a contextual cooperative reinforcement learning model to address the lack of cooperations between couriers while also considering system context \cite{Li2019}. Weirong et al. proposed a cooperative reinforcement learning algorithm for distributed economic dispatch in microgrids \cite{8306311}. Similar to previous studies, this paper aims to address the agent coordination issue in the box-pushing game.

In the box pushing game, agents collaborate to push a box towards the goal while avoiding obstacles. The minimalist agents have a small set of possible actions. They also choose their action based on limited observation capacities. The agents can only observe the environment using a box sensor placed in the middle of the box that is being pushed. The sensor gives limited environmental information such as the direction of the goal and nearby obstacles (see Sec.~\ref{subsec:action-state-space}). The agents are unable to sense one another -- all agents are unaware of one another. They can only act through their learned experience and their observation of the current environment. Because of this fact, agents often push against one other, especially towards the beginning of training. This leads to undesirable exploration. 

In this paper, we introduce the idea of a shared pool of information (SPI), which provides additional information generated and given to all agents at initialization. This idea is similar to common knowledge, in which agents use deduction as the basis for their action \cite{deWitt2019}. SPI aims to make up for a lack of communication between the agents by providing a common framework for all agents to base their exploration on. To generate an SPI that would enable such collaboration, it must pass a fitness test, which consists of an origin avoidance test and an angular spread test. The goal of the tests is to maximize total box displacement and the box's range of motion (see Sec~\ref{subsec:fitness}). We tested the efficiency of training in various challenging environments to observe the difference between SPI and random exploration.

Using the box-pushing game, this paper demonstrates how SPI is able to improve the ability for agents to coordinate during training. Furthermore, it adds minimal computation and overhead. In the experiment, SPI reduces instances in which agents cancel out one another’s actions, making each agents’ exploration more meaningful. This idea can be applied and used in tandem with other algorithms to help address the issue of coordination and cooperation for multi-agent problems in reinforcement learning.

\section{Related Works}

\subsection{The Box pushing problem}

The box pushing problem has been experimented on with single-agent and multi-agent RL. Studies find that both algorithms are equally as effective in simple environments. However, single agent RL proves to be more efficient and consistent in complex environments (\cite{4058979}, \cite{Rahimi2019}). In route planning, Ezra, José, and Gregorio proposed a new path planning algorithm that allows for the minimum number of reconfigurations, which can aid robots in planning and optimizing routes for the box pushing problem \cite{5342001}. For the box pushing problem, researchers also devised RL algorithms that are more efficient than the standard q-learning algorithm used for multi-agent RL (\cite{6899476}, \cite{6722515}). Toshiyuki, Kazuhiro, and Kazuaki proposed a technique that aims to organize what agents learn into clusters that represent the environment or situations that they might encounter. Each cluster contains its own set of rules to dictate what actions are optimal for agents in that specific scenario/environment \cite{6722515}. Kao-Shing, J.L., and Wei-Han proposed adaptive state aggregation q-learning that aids coordination by splitting the state space into chunks \cite{6899476}. Yang and Clarence used genetic algorithms alongside reinforcement learning, with an arbitrator evaluating the output of both to determine the agent's actions \cite{1470156}.

\subsection{Multiagent Communication}

Many multiagent reinforcement learning tasks require agents to coordinate in order to accomplish. In Markov Decision Process each "action" is a joint action for all agents and the reward is their combined reward. However, this creates a large action space \cite{Bellman1957}. Local optimization for the different agents via reward and value sharing (\cite{6224937}, \cite{principles}), direct policy search \cite{10.5555/2073946.2074003}, and Factored MDPs (\cite{10.5555/2980539.2980737}, \cite{1374220}) are approaches to address this problem. Agents can also observe one another’s actions and efforts implicitly rather than communicating directly to determine their own actions (\cite{1}, \cite{1641893}). However, implicit communication protocols for complex coordination problems typically require multiple training sessions for different components, severely increasing the training time and training complexity (\cite{10.5555/3305890.3306047}, \cite{implicit}). Sheng et al. developed coordinate graphs that are automatically and dynamically created for the agents to share information and decide on actions together \cite{li2021deepimplicitcoordinationgraphs}.

\subsection{Centralized vs Decentralized Learning}

Centralized training for decentralized execution (CLDE) is an approach to address the issue of coordination \cite{amato2024introductioncentralizedtrainingdecentralized}. Under the CLDE framework, Wenzhe et al. attempts to address the issue of coordination using f-MAT, which enables efficient collaboration through graph modeling \cite{fan2024efficientcollaborationgraphmodeling}. Centralized learning is useful to give agents information during training that they normally wouldn’t have access to during execution \cite{wang2023centralizedtrainingdecentralizedexecution}. There are two main ways of accomplishing this - a centralized critic for all agents or a centralized critic for each agent (\cite{lowe2020multiagentactorcriticmixedcooperativecompetitive}, \cite{10.5555/3504035.3504398}). Daewoo et al. proposed SchedNet where the centralized critic gives feedback which consists of message encoders, action selectors, and weight generators to each agent \cite{kim2019learningschedulecommunicationmultiagent}. Gang proposed Centralized Training and Exploration with Decentralized Execution via policy Distillation as an example of a critic for each agent \cite{10.5555/3398761.3398987}. In addition, Alexander et al. proposed a Bayesian mode to further optimize the exploration phase of CLDE training. It weighs the costs of exploration against its expected benefit gained from exploring \cite{10.5}. Decentralized learning is generally more computationally demanding or can encounter problems. But, in the real world, it is often impractical for a centralized model to access all agent’s actions and observations \cite{Shin2023}. In DL, nodes communicate and share models with their immediate neighbors, converging to a global model through local interactions \cite{pmlr-v119-koloskova20a}. CHOCO-SGD proposed by Anastasia et al. allows the communication and gradient computation steps to run in parallel \cite{Koloskova*2020Decentralized}. Although less research has been done on DL compared CL, Akash et al. created the DecentralizePy, hoping that others can use it to develop new DL systems \cite{10.1145/3578356.3592587}.

\section{Methodology}

In a simulated environment designed for agents to collaborate on the box-pushing task, agents rely on an implicit coordination framework called the Shared Pool of Information (SPI), which consists of a map and key. This framework provides agents with a shared reference that fosters aligned movements and minimizes counteractive actions. It aims to address the inefficiencies of random exploration in multi-agent reinforcement learning.

\subsection{Environment}

\begin{figure}[htbp]
    \centering
    \includegraphics[width=\textwidth]{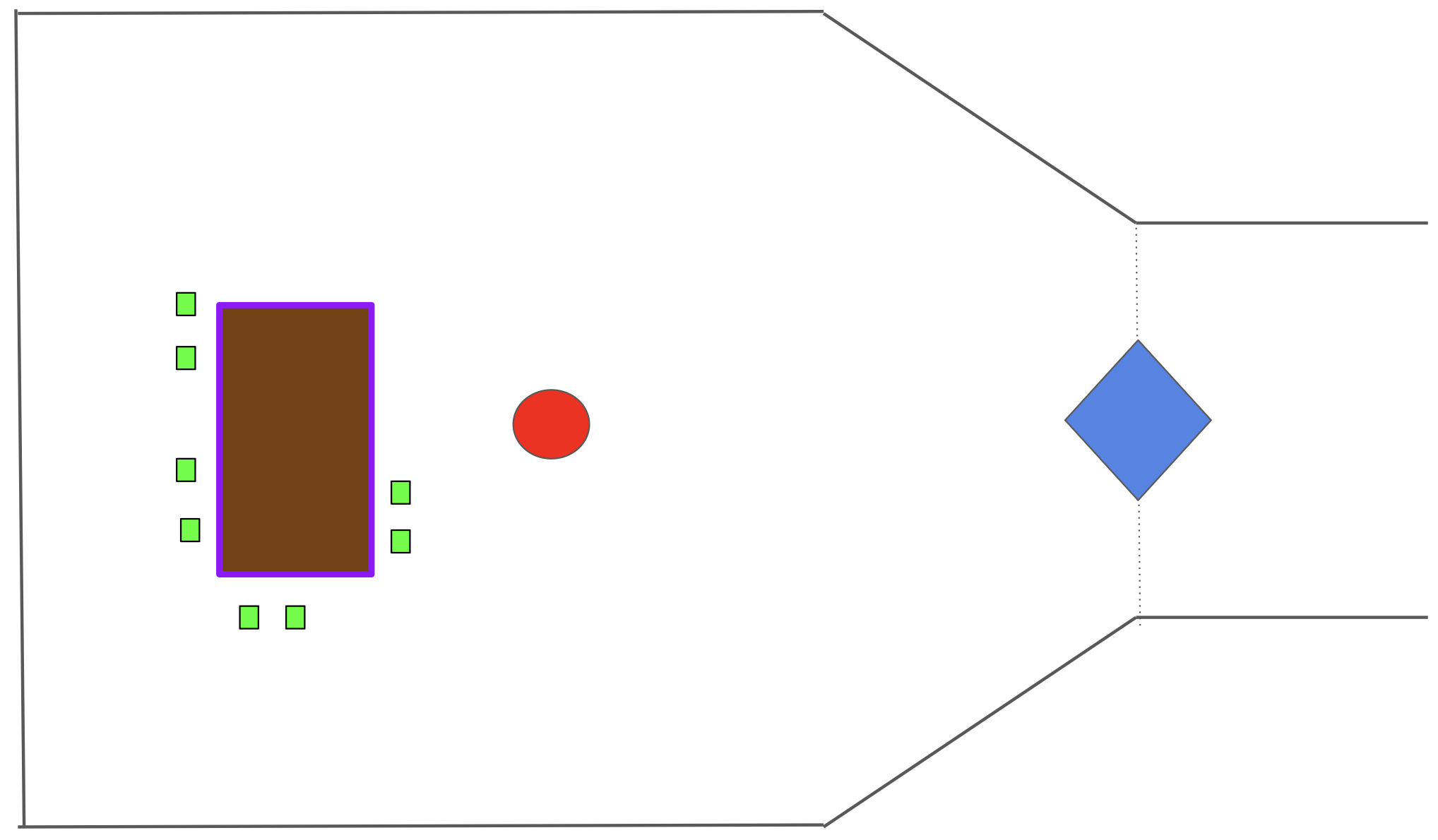}
    \caption{\textbf{Illustration of the box environment.}}
    \label{fig:env}
\end{figure}

Fig.~\ref{fig:env} shows the environment, which consists of patches, obstacles, agents, boxes, and the goal. The environment is built using the Pygame library. The graphical illustration demonstrates a possible state of the box-pushing task in which agents (green boxes) push the box (black box). Agents navigate around obstacles (represented by red circle) and walls (represented by black lines) to reach the goal (represented by blue diamond). Success occurs when the outer perimeter of the box collides with any area of the goal and failure happens when the outer perimeter of the box touches the obstacle or any wall. Failure can also happen when the box is unable to reach the goal in under 300 steps.

\subsection{Action Space and State Space}
\label{subsec:action-state-space}

\begin{figure}[htbp]
    \centering
    \begin{minipage}{0.48\textwidth}
        \centering
        \includegraphics[width=\textwidth]{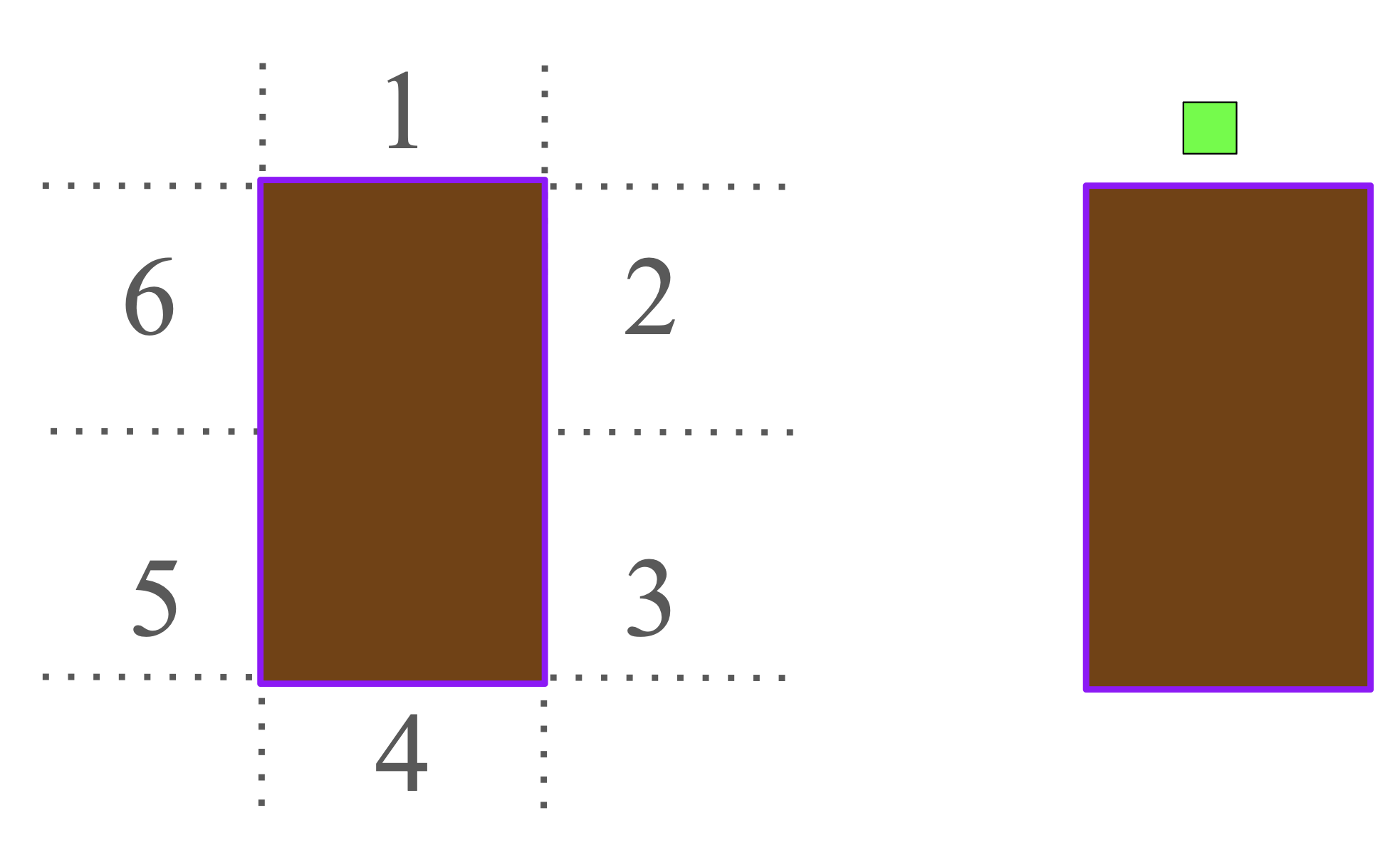} 
        \caption{\textbf{Illustration of the action space.}}
        \label{fig:as}
    \end{minipage}%
    \hfill 
    \begin{minipage}{0.48\textwidth}
        \centering
        \includegraphics[width=\textwidth]{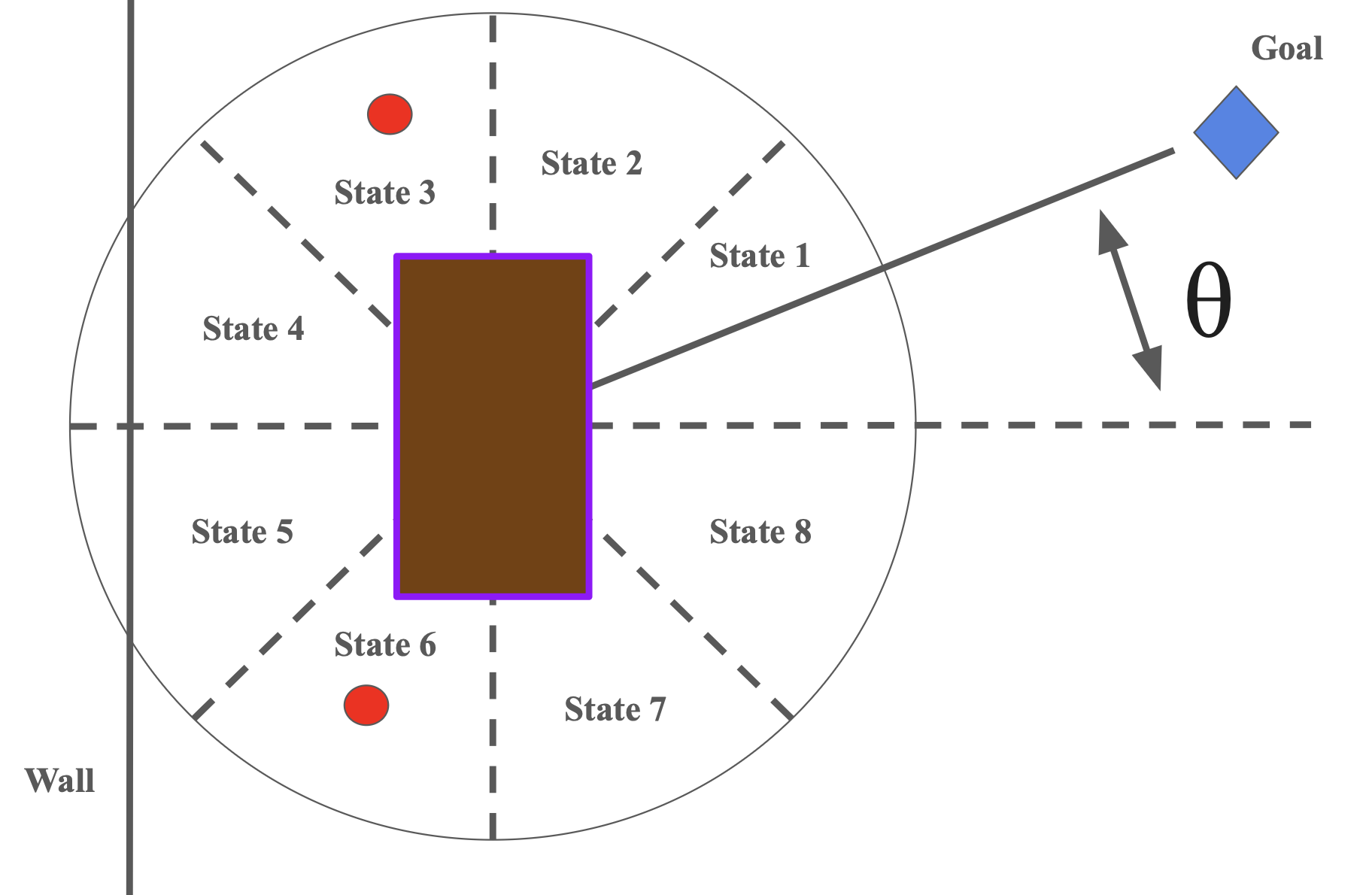} 
        \caption{\textbf{Illustration of the state space.}}
        \label{fig:ss}
    \end{minipage}
\end{figure}

At each time step, each agent chooses one region within the six available regions in the box neighborhood to apply force (see Fig.~\ref{fig:as}), directing the box’s movement accordingly. This information is subsequently converted into the action space \( A \), represented as follows:

\[
A = \{a_1, a_2, a_3, a_4, a_5, a_6\}
\]

In the neighborhood surrounding the box, six distinct regions are represented as \( a_1, a_2, \dots, a_6 \) (see Fig.~\ref{fig:as}). For each region, a value of 1 indicates that the agent has selected that region, while a value of 0 indicates that it has not. Each agent chooses exactly one region; therefore, \(\{1, 1, 0, 0, 0, 0\}\) and \(\{0, 0, 0, 0, 0, 0\}\) are not valid action spaces. In fig.~\ref{fig:as}, the agent selects region 1 of the box neighborhood. Therefore, the action space \( A \) is defined as \(\{1, 0, 0, 0, 0, 0\}\).

At each time step, the box’s sensor detects relevant environmental information within a 150-pixel radius as well as the angle between the box and the goal. This information is then converted into the state space \( S \), represented as follows:

\[
S = \{s_1, s_2, s_3, s_4, s_5, s_6, s_7, s_8, s_9\}
\]

\( s_1, s_2, \dots, s_8 \) represent the eight octants in fig.~\ref{fig:ss}. A value of 1 indicates the presence of an obstacle or wall in the octant, whereas a value of 0 indicates its absence. \( s_9 \) represents the angle between the box and the goal, scaled into the range \([-1, 1]\). In fig.~\ref{fig:ss}, an obstacle is present in states \( s_3 \) and \( s_6 \), and walls are present in states \( s_4 \) and \( s_5 \). Consequently, \( s_1, s_2, s_7 \), and \( s_8 = 0 \), while \( s_3, s_4, s_5 \), and \( s_6 = 1 \). The angle \( \theta = 30^\circ \) translates to \( s_9 = \frac{\theta - 180}{180} = -0.83 \). This gives the state space \( S = \{0, 0, 1, 1, 1, 1, 0, 0, -0.83\} \).

\subsection{Reward}

\textbf{Distance reward} (\(R_{\text{dis}}\)) is the reward for moving the box closer to the goal. \(D_{\text{old}}\) and \(D_{\text{new}}\) represent the distance in pixels between the box's sensor and the goal at the previous and current timesteps, respectively.

\[
R_{\text{dis}} = (D_{\text{old}} - D_{\text{new}}) \cdot 2.5
\]

\textbf{Rotation reward} (\(R_{\text{rot}}\)) is the reward designed to discourage excessive rotation of the box. \(\alpha_{\text{old}}\) and \(\alpha_{\text{new}}\) represent the angles between the box's x-axis and the goal at the previous and current timesteps, respectively.

\[
R_{\text{rot}} = \cos(\alpha_{\text{new}} - \alpha_{\text{old}}) - 0.98
\]

\textbf{Collision reward}, (\(R_{\text{col}}\)), is the negative reward applied if the box collides with an obstacle or wall.

\[
R_{\text{col}} = 
\begin{cases} 
0 & \text{if there is no collision} \\
-900 & \text{if there is a collision}
\end{cases}
\]

\textbf{Goal reward}, (\(R_{\text{goal}}\)), is the positive reward assigned when the box successfully reaches the goal.

\[
R_{\text{goal}} = 
\begin{cases} 
900 & \text{if goal is reached} \\
0 & \text{if goal is not reached}
\end{cases}
\]

The total reward is formulated as a weighted sum of distance, rotation, collision, and goal rewards. All reward weights are set to fixed values, with the distance reward assigned the highest weight. The values are defined in \cite{hao}.

\[
R_{\text{Total}} = w_1 R_{\text{dis}} + w_2 R_{\text{rot}} + w_3 R_{\text{col}} + w_4 R_{\text{goal}}
\]

\subsection{Speed Factor}

Speed factor controls how much force agents are able to exert. This impacts how much the agents can both displace and rotate the box. In this experiment, the speed factor is reduced to one-third and one-half of its original value, effectively limiting the agents' pushing strength. With reduced force, individual agents can no longer move the box as efficiently on their own. Cooperation becomes even more critical as failure to do so results in slower progress. This shift in dynamics is designed to test how the agents adapt to a more challenging environment, where collaborative effort is essential to achieving the desired outcome.

\subsection{Implicit Coordination Through Shared Pool of Information}

The primary goal of the introduction of a shared pool of information (SPI) is to encourage implied coordination -- the ability for agents to work constructively without specifically needing to exchange information between one another. In order to accomplish this, the scenario must have the following key characteristics:

\begin{enumerate}
\item \label{item:1} All agents have access to the same shared information at instantiation.
\item \label{item:2} Agents can only observe or read from the shared information.
\item \label{item:3} The shared information cannot be changed by the agents.
\item \label{item:4} The shared information is simple - needs little to no training and should have a small time and space complexity.
\end{enumerate}

The following ruleset ensures that agents, at instantiation, must be identical (\ref{item:1}) and cannot communicate with one another in any way by any means (\ref{item:2}, \ref{item:3}). The addition of SPI is efficient and lightweight -- adds no significant overhead (\ref{item:4}).

The inclusion of SPI reduces the likelihood that agents push the box from different sides, resulting in minimal movement. SPI should act like a silent guide that allows the agents to better align their efforts in order to improve the speed and efficiency of training.

\subsection{Proposed Shared Pool of Information}

\begin{figure}[htbp]
    \centering
    \includegraphics[width=\textwidth]{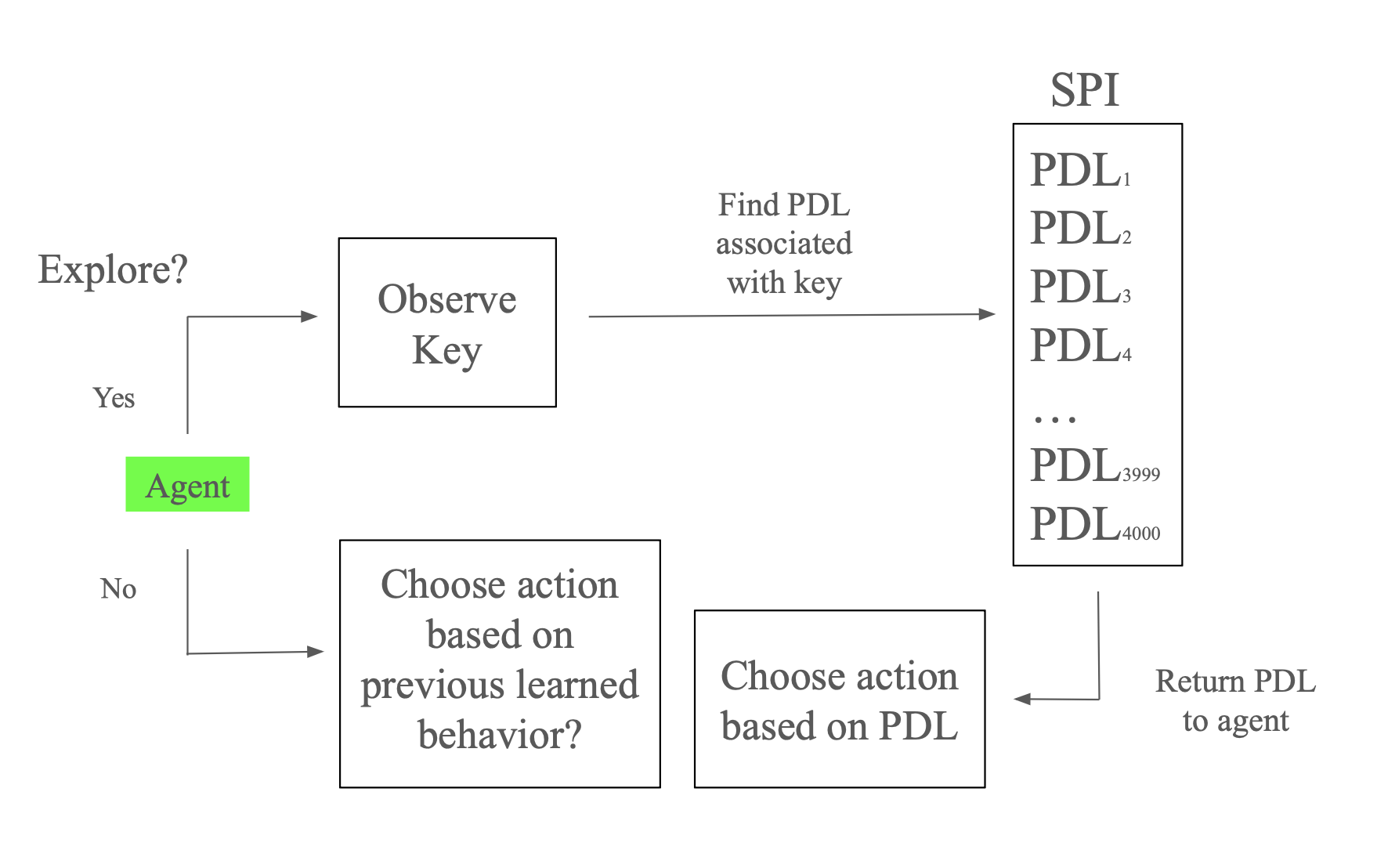}
    \caption{\textbf{This figure illustrates the procedure by which agents select actions during training.}}
    \label{fig:process}
\end{figure}

Instead of agents randomly exploring, the agents would now pseudo-randomly explore based on shared information. This is to improve the agents’ cooperation during training. The proposed SPI has two portions, map and key. The \textbf{map} refers to an idealized distribution outlining the potential outcomes and distribution of those outcomes when all the agents push the box simultaneously. In other words, we want to simulate the movement of the box under the assumption that the agents act as one or are controlled by a centralized entity. It is important to note that simply having such a distribution and enacting on its implementation would change the experiment from using multiagent reinforced learning (RL) approach to a single centralized agent RL approach by definition. Therefore, it is crucial to include a key, in which all agents can use to decipher the distribution, allowing for cooperation without the need for communication between agents.

The \textbf{key} must be able to be randomized and the randomization must have sufficient range -- larger than the number of agents in the experiment. The distribution of the randomized key must also be uniform to ensure no agents will have no bias to any distribution outlined in the map. Although there are numerous possible implementations of the key, in the experiment, it is simply generated with a random number generator using the random library in Python. The key will be generated every step of training and all agents will be able to observe and read from it.

The map of SPI contains a large number of probability distribution lists (PDL) -- the exact number is discussed in section~\ref{sec:numPDL}.  As there are six different actions the agents can take, each PDL will have six entries, representing the probability that an agent should choose each action. The summation of the entries of each PDL should be exactly 1. Examples of valid and invalid PDLs are in Table~\ref{tab:pdl_examples}.

During exploration, agents use the map by first reading the generated key. Then, the agent must find the PDL associated with the key. Given the PDL, the agent will subsequently use it to determine what action to take. Fig.~\ref{fig:process} is a drawn example of the process.

\renewcommand{\arraystretch}{1.75} 

\begin{table}[htbp]
    \centering
    \begin{tabular}{|c|>{\centering\arraybackslash}m{9cm}|}
        \hline
        \textbf{PDL Type} & \textbf{Examples} \\
        \hline
        Valid PDL & $\left[ \frac{1}{6}, \frac{1}{6}, \frac{1}{6}, \frac{1}{6}, \frac{1}{6}, \frac{1}{6} \right]$, 
                    $\left[ 1, 0, 0, 0, 0, 0 \right]$, 
                    $\left[ \frac{1}{6}, \frac{1}{6}, \frac{1}{4}, \frac{1}{4}, \frac{1}{12}, \frac{1}{12} \right]$ \\
        \hline
        Invalid PDL & $\left[ \frac{1}{7}, \frac{1}{7}, \frac{1}{7}, \frac{1}{7}, \frac{1}{7}, \frac{1}{7} \right]$, 
                      $\left[ \frac{1}{5}, \frac{1}{5}, \frac{1}{5}, \frac{1}{5}, \frac{1}{5}, \frac{1}{5} \right]$, 
                      $\left[ \frac{1}{4}, \frac{1}{4}, \frac{1}{4}, \frac{1}{4} \right]$ \\
        \hline
    \end{tabular}
    \caption{Examples of Valid and Invalid PDL}
    \label{tab:pdl_examples}
\end{table}

\subsection{Fitness Test using BMD}
\label{subsec:fitness}

At each step, agents will simultaneously push the box after determining what action to take. In one simulation, the displacement of the box in both the x and y axis in which the box is pushed from its starting point is recorded. Each simulation is exactly one step long. After a significant amount of simulations, we can plot out the resulting displacement distribution of the box. We call that box movement distribution (BMD). Figure 4 shows an example of such a graph.
When using the map, agents should be able to collaborate efficiently. To be considered effective, the map needs to pass a fitness test, which consists of an origin avoidance test and a uniformity of angular spread test. We must ensure that BMD minimizes the amount of conflicting forces exerted by the agents while allowing agents to push the box in all directions with equal likelihoods. 

Figure 4: Graph

\subsubsection{Origin Avoidance}

The first component of the fitness test evaluates origin avoidance, which measures the percentage of time the agents can displace the box a sufficient distance. The origin avoidance fitness score is calculated as follows:

\[
\text{origin\_avoidance\_fitness} = 1 - \left( \frac{\text{near\_origin\_count}}{\text{total\_count}} \right)
\]

In the tests, the box is considered near the origin if its total displacement is three times less than its maximum potential displacement. This ensures the agents cooperate so that the box is sufficiently displaced from its original position. The count of instances where the box remains near the origin is recorded as \textbf{near\_origin\_count}, while \textbf{total\_count} denotes the overall count of test instances. A low origin avoidance fitness score would indicate that the agents constantly exert conflicting forces, unable to consistently displace the box far from the origin. On the other hand, a high origin avoidance fitness score would indicate that the agents are generally able to collaborate, as they rarely exert conflicting forces.

\subsubsection{Uniformity of Angular Spread}

The second component of the fitness test evaluates the uniformity of the box's angular spread, which tests if the box can be pushed in any direction with equal likelihood. To calculate the uniformity of angular spread, the angle of each box from the origin is computed. Then, we define bin edges for \(n\) equally sized bins, each representing \( \frac{360}{n} \) degrees, where \(n\) is the number of agents. Next, we compute the histogram of the angles (see fig.~\ref{fig:comparison}) and utilize the mean and standard deviation of the histogram to determine the coefficient of variation (CV). A lower CV indicates higher uniformity in angular distribution. Therefore, we assess the uniformity of the angular spread by inverting the CV and normalizing it to a \([0, 1]\) range, as shown in the following equation:

\[
\text{uniformity\_of\_angular\_spread} = \left( \frac{1 - \text{CV}}{\text{CV}} \right)
\]

A low fitness score indicates insufficient diversification in the box’s movement directions, which can hinder the training process. Conversely, a higher score reflects that, throughout the simulations, the agents have successfully aligned themselves to enable multi-directional pushing of the box. Such flexibility in angular spread is essential for maximizing the range of the box’s potential movement.

\section{Results and Discussion}

\subsection{Generating PDL}

There are 6 potential actions for each agent (see fig.~\ref{fig:as}). In terms of box displacement, action 2 and 3 along with action 5 and 6 are effectively the same. For calculating the ideal distribution, we can simplify the state actions to the 4 actions shown. The instruction to generate one PDL is as follows.

\subsubsection{Generating the Initial Values}
First generates a random float, \( \texttt{value1} \), from a uniform distribution in range \([0, \textbf{cap}]\). This will serve as the 0th element in the first PDL:

\[
\texttt{value1} \sim \mathcal{U}(0, cap)
\]

The remaining sum represents the total the remaining three elements need to sum up to:

\[
\texttt{remaining\_sum} = 1 - \texttt{value1}
\]

Then, three additional random values are generated.

\[
\texttt{value2} \sim \mathcal{U}(0, 1), \quad \texttt{value3} \sim \mathcal{U}(0, 1), \quad \texttt{value4} \sim \mathcal{U}(0, 1)
\]

The three generated values are normalized to sum to 1. Then, these values are scaled by \texttt{remaining\_sum} to ensure that the total sum of all three elements equals 1:

\[
\begin{array}{ll}
\texttt{value2}_{\text{norm}} = \frac{\texttt{value2}}{\texttt{value2} + \texttt{value3} + \texttt{value4}} &
\texttt{value2}_{\text{scaled}} = \texttt{value2}_{\text{norm}} \times \texttt{remaining\_sum} \\
\texttt{value3}_{\text{norm}} = \frac{\texttt{value3}}{\texttt{value2} + \texttt{value3} + \texttt{value4}} &
\texttt{value3}_{\text{scaled}} = \texttt{value3}_{\text{norm}} \times \texttt{remaining\_sum} \\
\texttt{value4}_{\text{norm}} = \frac{\texttt{value4}}{\texttt{value2} + \texttt{value3} + \texttt{value4}} &
\texttt{value4}_{\text{scaled}} = \texttt{value4}_{\text{norm}} \times \texttt{remaining\_sum}
\end{array}
\]

The complete list of values, \texttt{all\_values}, is created by concatenating \texttt{value1} with the scaled values:

\[
\texttt{all\_values} = [\texttt{value1}, \texttt{value2}_{\text{scaled}}, \texttt{value3}_{\text{scaled}}, \texttt{value4}_{\text{scaled}}]
\]

\subsubsection{Condition Check for Validity}
The difference between \(\text{value1}\) and \(\texttt{value3}_{\text{scaled}}\) must be minimum \textbf{margin value}. If this condition is not met, the generation process restarts:

\[
|\texttt{value}_1 - \texttt{value3}_{\texttt{scaled}}| \geq margin
\]

\subsubsection{Generating the \texttt{PDL\_4x4} Matrix}
   The next step is to create a \(4 \times 4\) matrix, \texttt{PDL\_4x4}. Each row \( i \) is a cyclic permutation (shift) of \( \textbf{all\_values} \) by \( i \) positions:

\[
\textbf{PDL}_{4 \times 4} = 
\begin{bmatrix}
    \text{value}_1 & \texttt{value2}_{\text{scaled}} & \texttt{value3}_{\text{scaled}} & \texttt{value4}_{\text{scaled}} \\
    \texttt{value4}_{\text{scaled}} & \text{value}_1 & \texttt{value2}_{\text{scaled}} & \texttt{value3}_{\text{scaled}} \\
    \texttt{value3}_{\text{scaled}} & \texttt{value4}_{\text{scaled}} & \text{value}_1 & \texttt{value2}_{\text{scaled}} \\
    \texttt{value2}_{\text{scaled}} & \texttt{value3}_{\text{scaled}} & \texttt{value4}_{\text{scaled}} & \text{value}_1 \\
\end{bmatrix}
\]

In general, each element in row \( i \) and column \( j \) of \( \textbf{PDL}_{4 \times 4} \) is defined as:

\[
\textbf{PDL}_{4 \times 4}[i, j] = \text{value}_{((j + i) \bmod 4) + 1}
\]

\subsubsection{Generating the \texttt{PDL\_4x6} Matrix} 
   A \(4 \times 6\) matrix, \texttt{PDL\_4x6}, is created using specific transformations of \texttt{PDL\_4x4}. The columns of \texttt{PDL\_4x6} are populated as follows:
   
\begin{itemize}
    \item The 0th column of \texttt{PDL\_4x6} matches the 0th column of \texttt{PDL\_4x4}:
    \[
    \texttt{PDL\_4x6}[i, 0] = \texttt{PDL\_4x4}[i, 0].
    \]
    \item The 1st and 2nd columns are each half of the 1st column of \texttt{PDL\_4x4}:
    \[
    \texttt{PDL\_4x6}[i, 1] = \frac{1}{2} \cdot \texttt{PDL\_4x4}[i, 1], \quad \texttt{PDL\_4x6}[i, 2] = \frac{1}{2} \cdot \texttt{PDL\_4x4}[i, 1].
    \]
    \item The 3rd column matches the 2nd column of \texttt{PDL\_4x4}:
    \[
    \texttt{PDL\_4x6}[i, 3] = \texttt{PDL\_4x4}[i, 2].
    \]
    \item The 4th and 5th columns are each half of the 3rd column of \texttt{PDL\_4x4}:
    \[
    \texttt{PDL\_4x6}[i, 4] = \frac{1}{2} \cdot \texttt{PDL\_4x4}[i, 3], \quad \texttt{PDL\_4x6}[i, 5] = \frac{1}{2} \cdot \texttt{PDL\_4x4}[i, 3].
    \]
\end{itemize}

\subsubsection{Normalizing Rows in \texttt{PDL\_4x6}}

Each row in \texttt{PDL\_4x6} is normalized to ensure the values sum to 1.

\[
\texttt{row\_sums}[i] = \sum_{j} \texttt{PDL\_4x6}[i, j].
\]

\[
\texttt{PDL\_4x6}[i, j] = \frac{\texttt{PDL\_4x6}[i, j]}{\texttt{row\_sums}[i]}.
\]

Four valid PDLs are generated: \texttt{PDL\_4x6}

\subsubsection{Number of PDLs Generated} \label{sec:numPDL}

\begin{figure}[H]
    \centering
    \begin{minipage}[t]{0.48\textwidth}
        \centering
        \includegraphics[width=\textwidth]{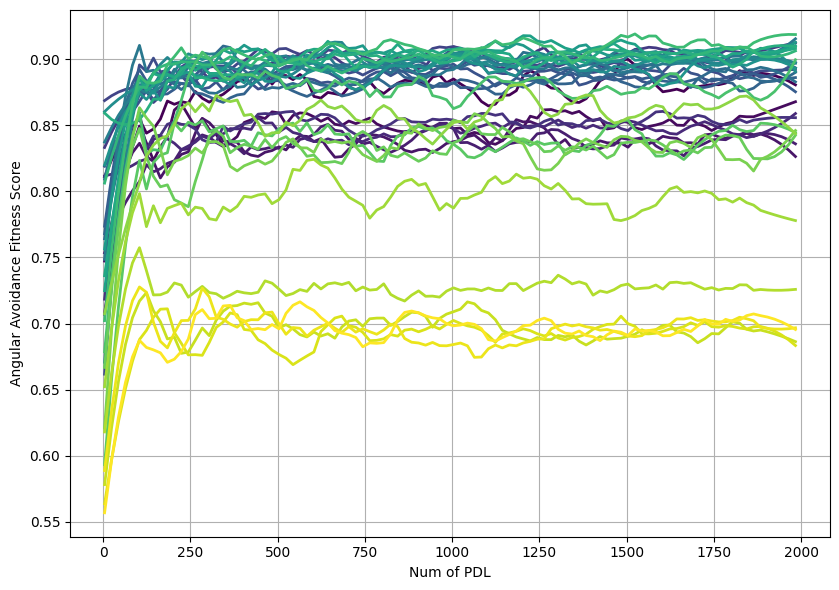}
        \caption{\textbf{The angular avoidance fitness score is tested for different margin and cap values. The lighter lines represent a high margin and cap value. The darker lines represent a low margin and cap value.}}
        \label{fig:numpdl}
    \end{minipage}
    \hfill
    \begin{minipage}[t]{0.48\textwidth}
        \centering
        \includegraphics[width=\textwidth]{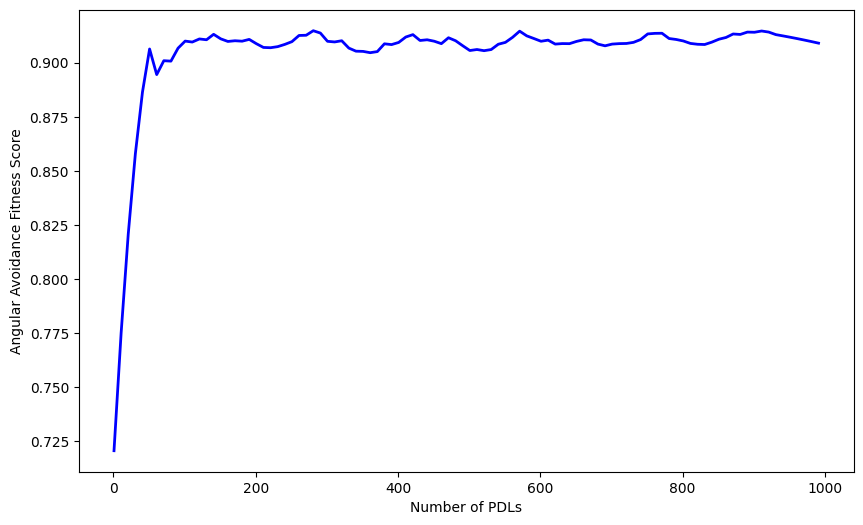}
        \caption{\textbf{The angular avoidance fitness score is calculated for different numbers of PDLs using a margin value of 0.3 and a cap value of 0.1.}}
        \label{fig:pdl15}
    \end{minipage}
\end{figure}

When there are 15 agents, the angular avoidance fitness score increases with more PDL distributions. After about 250 PDLs, additional PDLs seem to have little to no effect on the angular avoidance fitness score. This trend also holds true given any margin and cap values (see Fig.~\ref{fig:numpdl}). In the experiment, we will use 4000 PDLs to ensure a high angular avoidance fitness score.

\subsubsection{Generating Cap and Margin Values}

In the experiment, 15 agents and 4000 PDLs are used. The optimal margin value and cap value for those parameters are tested. 100000 total simulations are run and the results are plotted in a line graph. 

As margin value increases, the origin avoidance fitness score also increases. A smaller cap value leads to a higher origin avoidance fitness score. However, with a higher margin value (ie. 0.5), the significance of cap value decreases.

Until a margin value of 0.30, by increasing margin value, angular spread fitness score also increases. Afterwards, an increase in margin value drastically decreases the angular spread fitness score. Cap value does not have a meaningful impact on angular spread fitness score.

The cap chosen for the experiment is 0.1 while the margin value is 0.3. This set of values is chosen because it maximizes angular spread fitness score while maintaining a high origin avoidance fitness score. There exist many values of margin value and fitness value in which both origin avoidance and angular spread test are sufficiently passed. It is up to the experimenter to determine if origin avoidance is more valuable or angular spread.

\subsubsection{Fitness Test - New vs Random Explore}

\begin{figure}[htbp]
    \centering
    \begin{minipage}[b]{0.4\textwidth}
        \centering \textbf{SPI}
    \end{minipage}
    \begin{minipage}[b]{0.4\textwidth}
        \centering \textbf{Random Exploration}
    \end{minipage}
    
    \begin{minipage}[b]{0.4\textwidth}
        \centering
        \includegraphics[width=\textwidth]{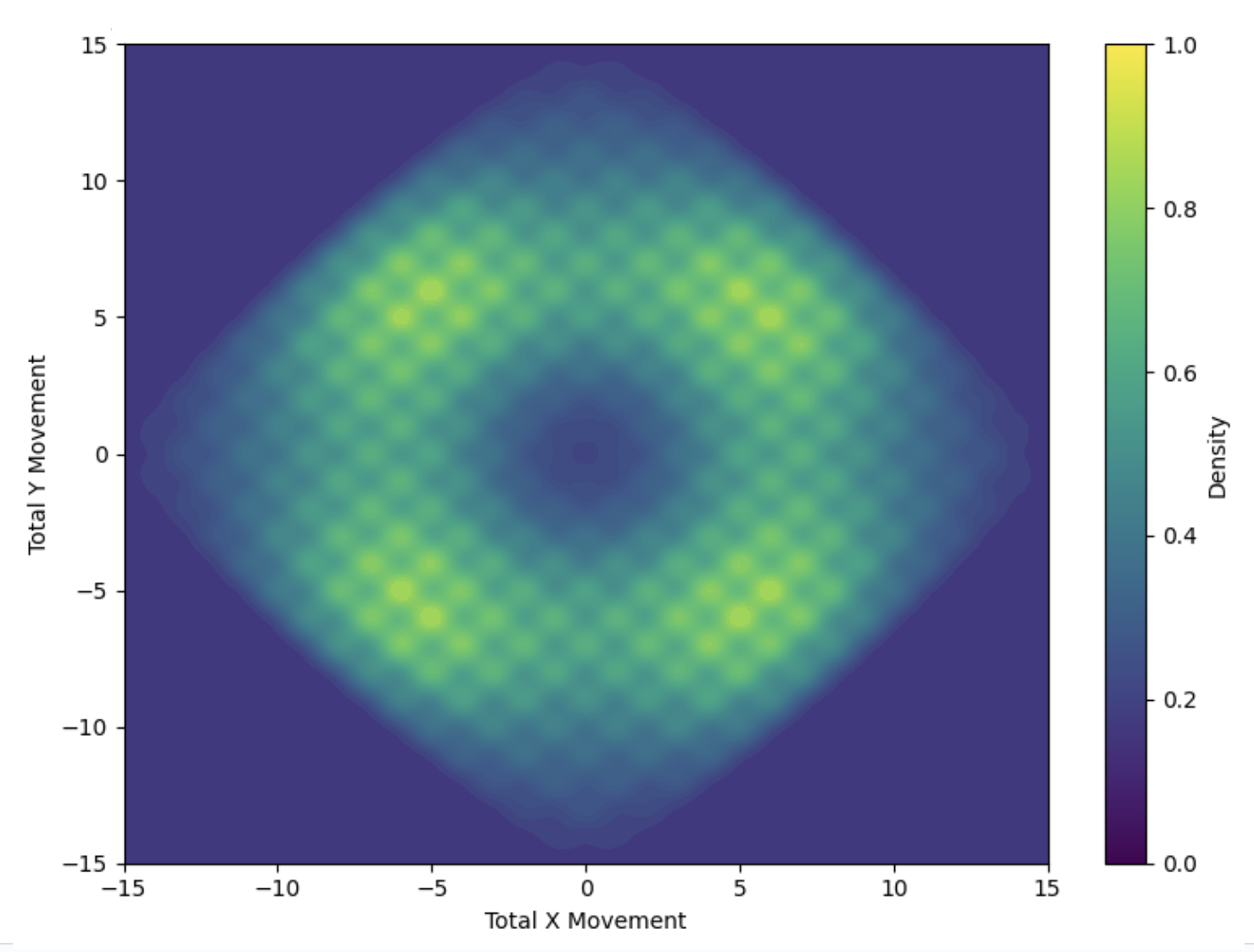}
    \end{minipage}
    \hspace{0.05\textwidth} 
    \begin{minipage}[b]{0.4\textwidth}
        \centering
        \includegraphics[width=\textwidth]{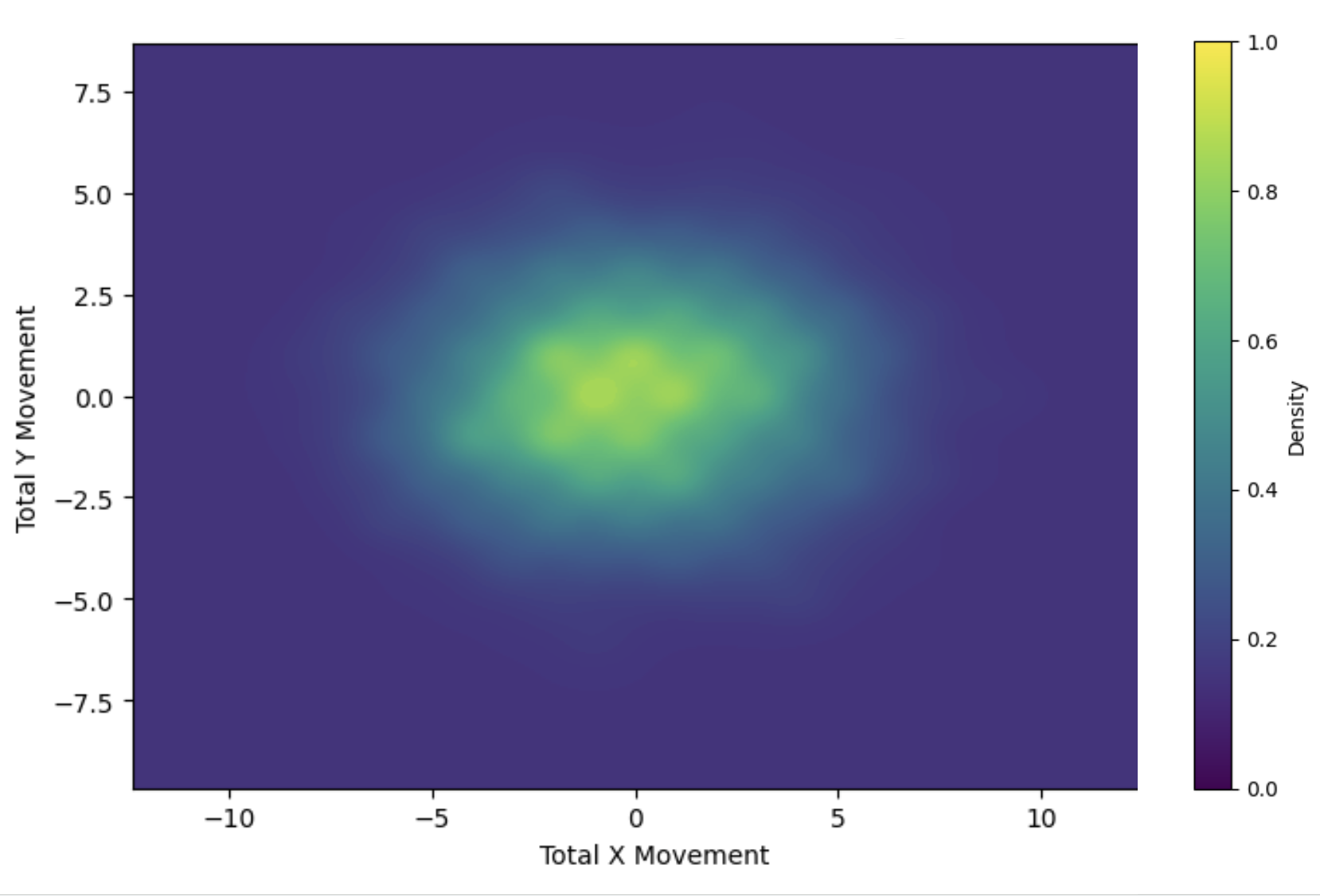}
    \end{minipage} \\
    \begin{minipage}[b]{\textwidth}
        \centering \small Kernel density estimate (KDE) plot of box movement from origin.
    \end{minipage}
    
    \vspace{1em} 
    \begin{minipage}[b]{0.4\textwidth}
        \centering
        \includegraphics[width=\textwidth]{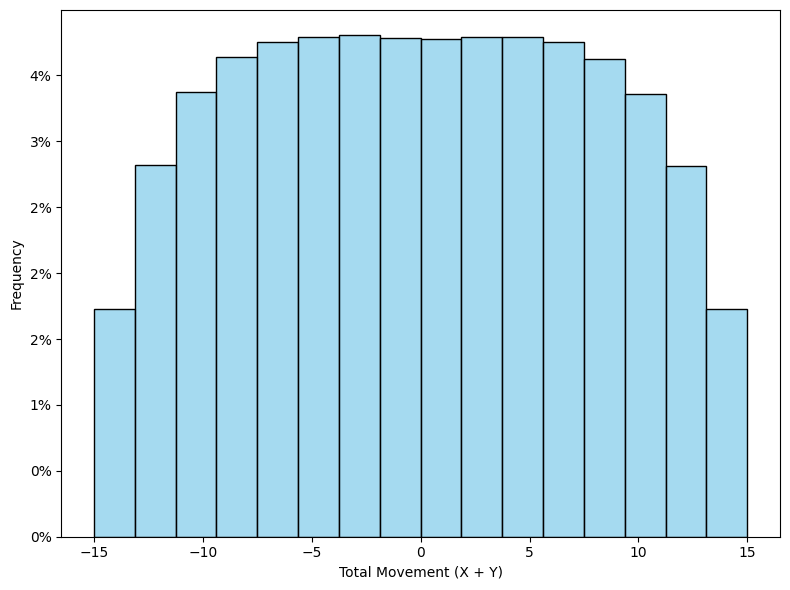}
    \end{minipage}
    \hspace{0.05\textwidth}
    \begin{minipage}[b]{0.4\textwidth}
        \centering
        \includegraphics[width=\textwidth]{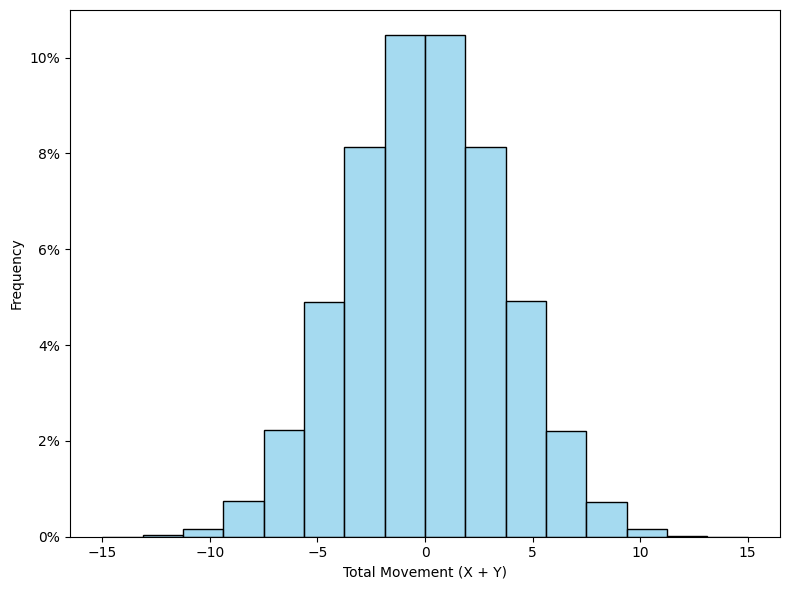}
    \end{minipage} \\
    \begin{minipage}[b]{\textwidth}
        \centering \small Histogram of box displacement.
    \end{minipage}
    
    \vspace{1em} 
    \begin{minipage}[b]{0.4\textwidth}
        \centering
        \includegraphics[width=\textwidth]{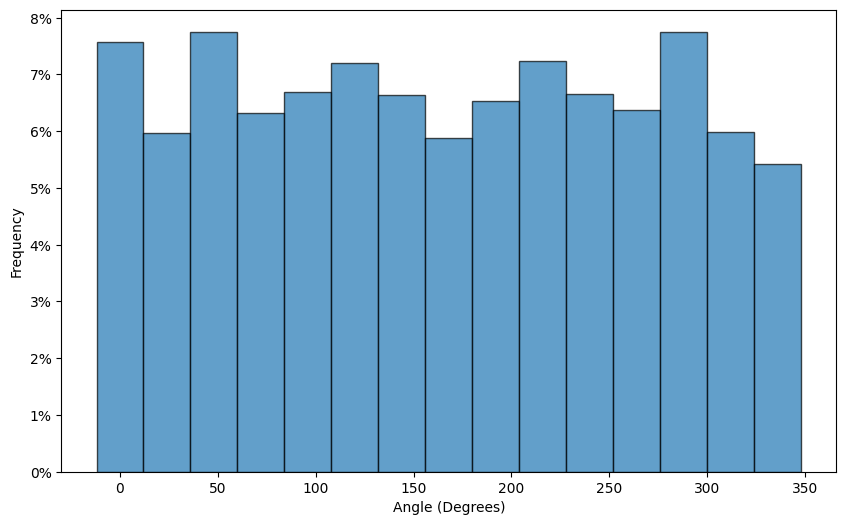}
    \end{minipage}
    \hspace{0.05\textwidth}
    \begin{minipage}[b]{0.4\textwidth}
        \centering
        \includegraphics[width=\textwidth]{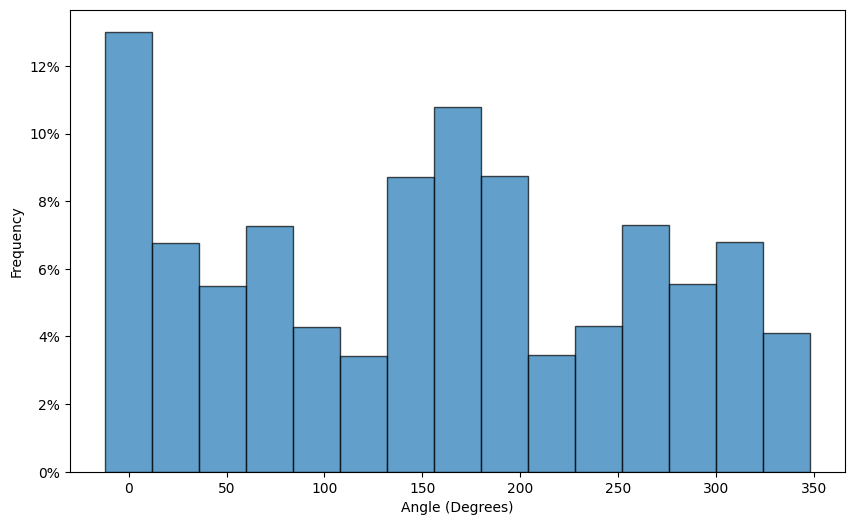}
    \end{minipage} \\
    \begin{minipage}[b]{\textwidth}
        \centering \small Histogram of directional box movement.
    \end{minipage}

    \caption{\textbf{Comparison of SPI and random exploration for stuffs}}
    \label{fig:comparison}
\end{figure}

The old method of random exploration has a origin avoidance score of 0.215 and an angular spread score of 0.714. The new method has a origin score of 0.899 and an angular spread score of 0.911. By comparing the differences between both methods using the box movement graph and histogram of point directions, it is evident that by using SPI, the box not only has access to a wider range of movement, but also can explore all directions more uniformly (see Fig.~\ref{fig:comparison}).

\subsection{Experiment Results}

Two experiments are conducted to test the effectiveness of SPI. Each experiment is run with a different speed factor (\(\frac{1}{2}\) and \(\frac{1}{3}\)). In each experiment, we plotted, compared, and evaluated the performance measures (step distance, runtime, reward, and success rate) of the training phase between random exploration and the proposed SPI model.

\subsubsection{Success Rate}

\begin{figure}[htbp]
    \centering
    \begin{subfigure}[t]{0.48\textwidth}
        \centering
        \includegraphics[width=\textwidth]{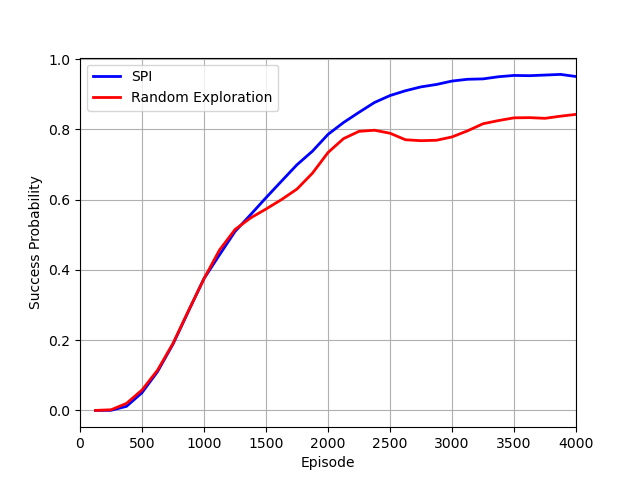}
        \caption{1/3 speed factor}
        \label{fig:suc3}
    \end{subfigure}
    \hfill
    \begin{subfigure}[t]{0.48\textwidth}
        \centering
        \includegraphics[width=\textwidth]{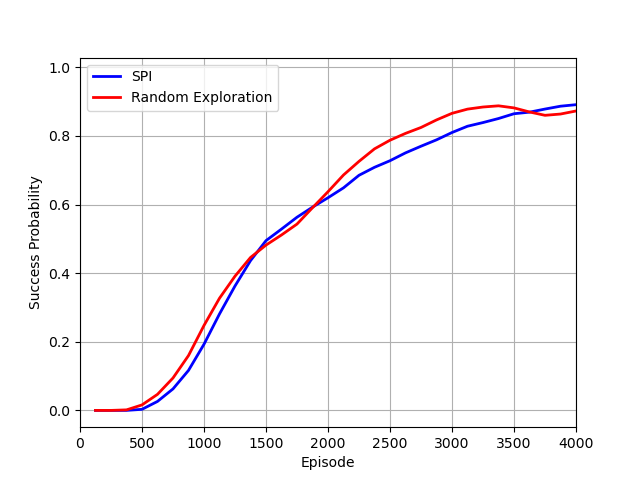}
        \caption{1/2 speed factor}
        \label{fig:suc2}
    \end{subfigure}
    \caption{\textbf{This plot illustrates the success rate of the box reaching the goal across training episodes.}}
    \label{fig:suc}
\end{figure}

Success is defined as the scenario in which the box reaches the goal without hitting an obstacle. Failure is defined as the scenario in which the box hits an obstacle. Success rate measures the frequency of success in each training episode. For both SPI and random exploration, the success rate typically increases throughout training. It rises quickly at first, but it plateaus once the success rate becomes high. The episodic success rate for both algorithms at the beginning of training is nearly identical. However, especially for a speed factor of \(\frac{1}{3}\), SPI has a higher success rate towards the end (see fig.~\ref{fig:suc}).

\subsubsection{Steps}

We define a \textbf{step} as every instance in which all agents select an action simultaneously. During each step, agents will collectively push the box, resulting in a single displacement of the box. As shown in Figures ?, the trend of step count for speed factors of \(\frac{1}{3}\) and \(\frac{1}{2}\) are very similar. However, there is a slightly larger discrepancy between SPI and random exploration for a speed factor of \(\frac{1}{3}\).

\begin{figure}[htbp]
    \centering
    \begin{subfigure}[t]{0.48\textwidth}
        \centering
        \includegraphics[width=\textwidth]{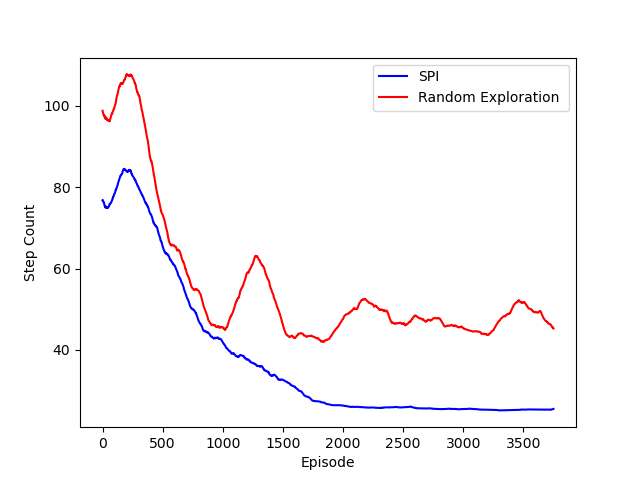}
        \caption{1/3 speed factor}
        \label{fig:step3}
    \end{subfigure}
    \hfill
    \begin{subfigure}[t]{0.48\textwidth}
        \centering
        \includegraphics[width=\textwidth]{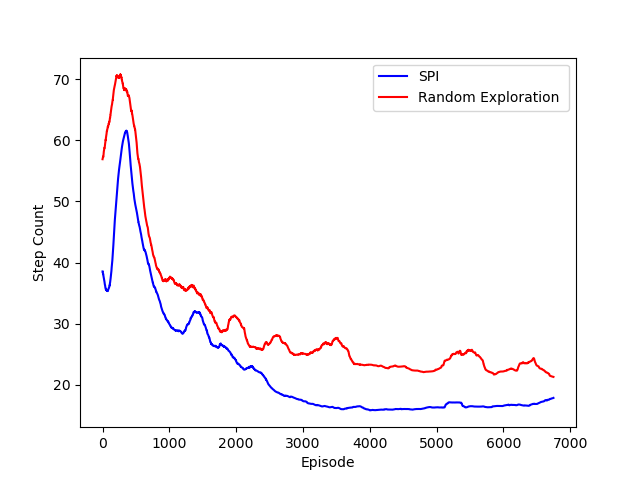}
        \caption{1/2 speed factor}
        \label{fig:step2}
    \end{subfigure}
    \caption{\textbf{These figures illustrates and compares the step count between SPI and random exploration at two different speed factors: 1/3 and 1/2.}}
    \label{fig:step}
\end{figure}

\begin{figure}[H]
    \centering
    \begin{minipage}[t]{\textwidth}
        \centering
        \makebox[0.48\textwidth]{\textbf{1/3 Speed Factor}}
        \hfill
        \makebox[0.48\textwidth]{\textbf{1/2 Speed Factor}}

        \begin{minipage}[t]{0.48\textwidth}
            \centering
            \includegraphics[width=\textwidth]{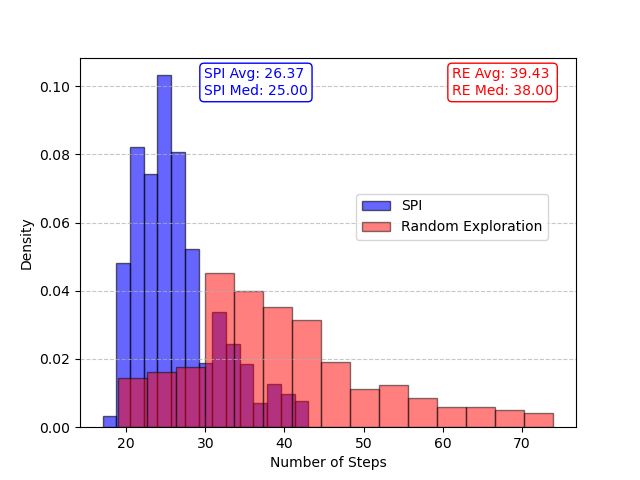}
        \end{minipage}
        \hfill
        \begin{minipage}[t]{0.48\textwidth}
            \centering
            \includegraphics[width=\textwidth]{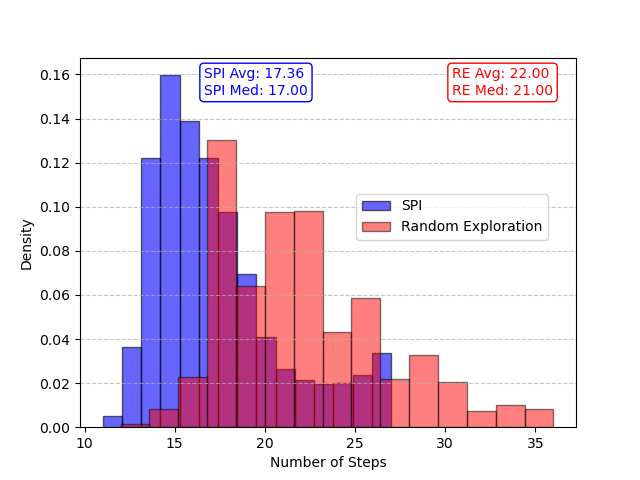}
        \end{minipage}

        \caption{\textbf{These figures illustrate and compare the success steps taken during all of training at two different speed factors: 1/3 and 1/2.}}
        \label{fig:stepsucall}
    \end{minipage}
\end{figure}

\begin{figure}[H]
    \centering
    \begin{minipage}[t]{\textwidth}
        \centering
        \makebox[0.48\textwidth]{\textbf{1/3 Speed Factor}}
        \hfill
        \makebox[0.48\textwidth]{\textbf{1/2 Speed Factor}}

        \begin{minipage}[t]{0.48\textwidth}
            \centering
            \includegraphics[width=\textwidth]{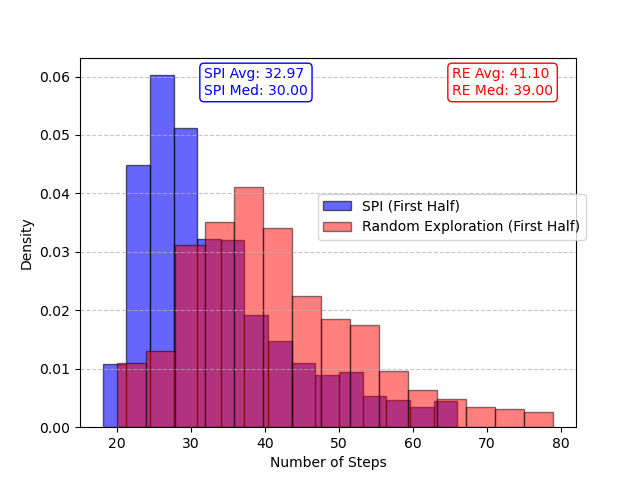}
        \end{minipage}
        \hfill
        \begin{minipage}[t]{0.48\textwidth}
            \centering
            \includegraphics[width=\textwidth]{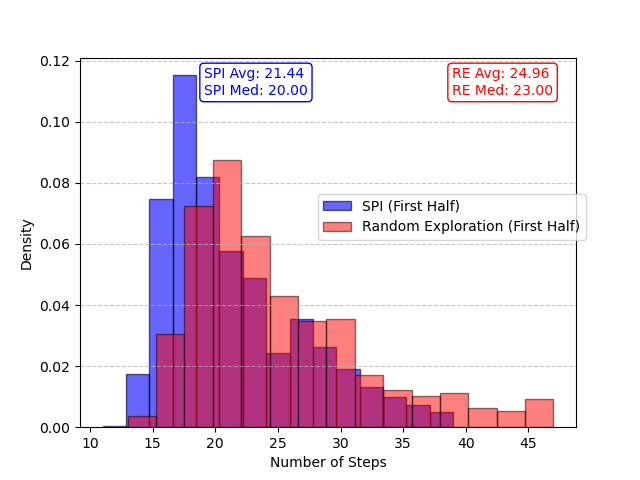}
        \end{minipage}

        \caption{\textbf{These figures illustrate and compare the success steps taken during the first half of training at two different speed factors: 1/3 and 1/2.}}
        \label{fig:stepsucfirst}
    \end{minipage}
\end{figure}

\begin{figure}[H]
    \centering
    \begin{minipage}[t]{\textwidth}
        \centering
        \makebox[0.48\textwidth]{\textbf{1/3 Speed Factor}}
        \hfill
        \makebox[0.48\textwidth]{\textbf{1/2 Speed Factor}}

        \begin{minipage}[t]{0.48\textwidth}
            \centering
            \includegraphics[width=\textwidth]{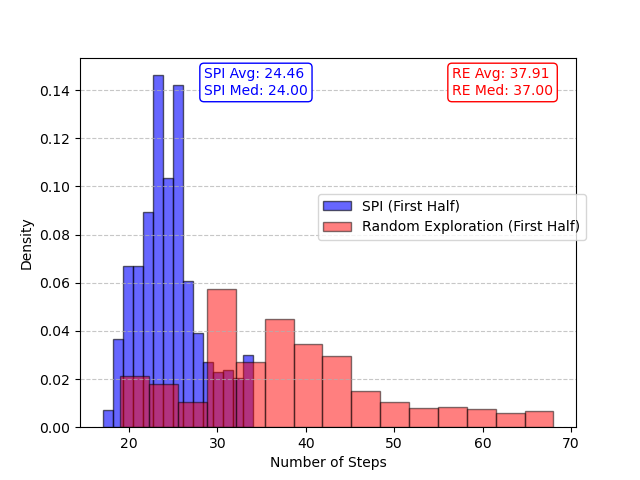}
        \end{minipage}
        \hfill
        \begin{minipage}[t]{0.48\textwidth}
            \centering
            \includegraphics[width=\textwidth]{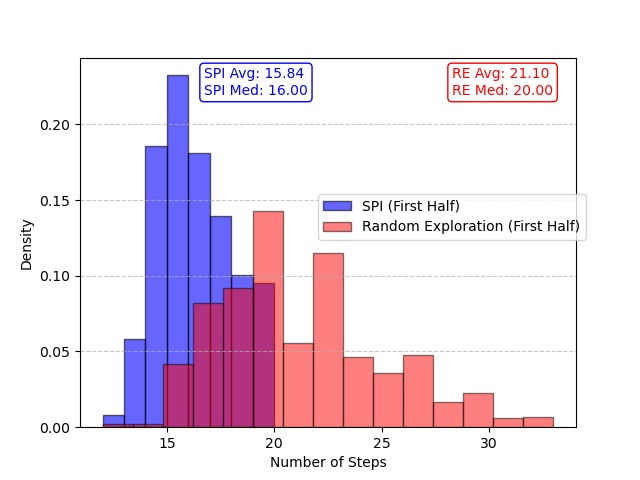}
        \end{minipage}

        \caption{\textbf{These figures illustrate and compare the success steps taken during the second half of training at two different speed factors: 1/3 and 1/2.}}
        \label{fig:stepsucsecond}
    \end{minipage}
\end{figure}

We define \textbf{success steps} and \textbf{failure steps} as the number of steps in a successful or failed training episode, respectively. The difference in step efficiency between SPI and random exploration can largely be attributed to the success steps (see Fig.~\ref{fig:stepsucall}). Early in training, agents using SPI are able to find more efficient successful routes than agents using random exploration (see Fig.~\ref{fig:stepsucfirst}). Towards the end of training, agents using SPI consistently use optimal routes to reach the goal, while agents using random exploration struggle to learn to reach the goal quickly (see Fig.~\ref{fig:stepsucsecond}).

\begin{figure}[H]
    \centering
    \begin{minipage}[t]{\textwidth}
        \centering
        \makebox[0.48\textwidth]{\textbf{1/3 Speed Factor}}
        \hfill
        \makebox[0.48\textwidth]{\textbf{1/2 Speed Factor}}

        \begin{minipage}[t]{0.48\textwidth}
            \centering
            \includegraphics[width=\textwidth]{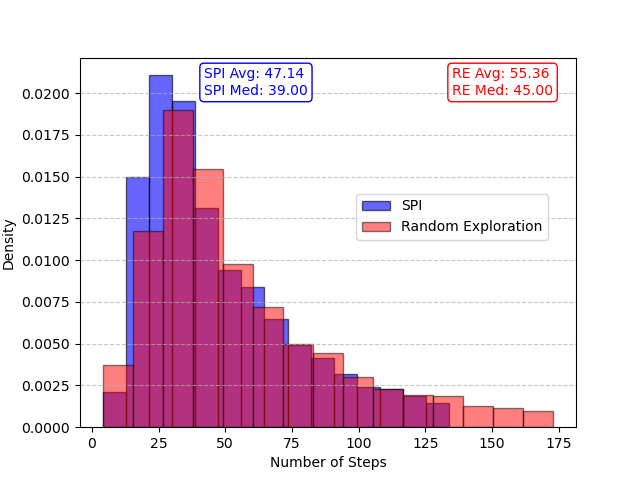}
        \end{minipage}
        \hfill
        \begin{minipage}[t]{0.48\textwidth}
            \centering
            \includegraphics[width=\textwidth]{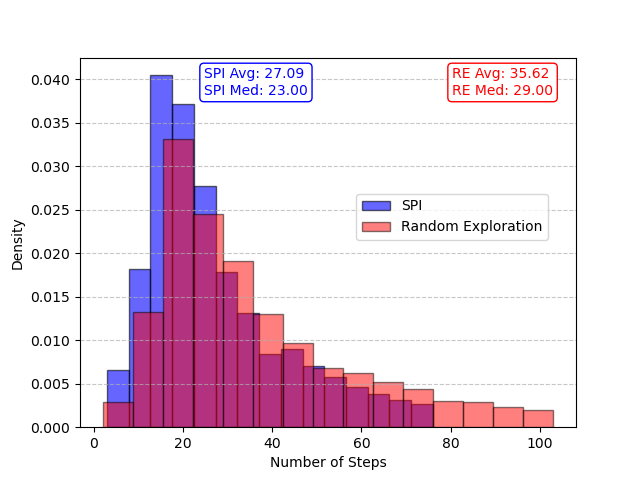}
        \end{minipage}

        \caption{\textbf{These figures illustrate and compare the failure steps taken during all of training at two different speed factors: 1/3 and 1/2.}}
        \label{fig:stepfailall}
    \end{minipage}
\end{figure}

\begin{figure}[H]
    \centering
    \begin{minipage}[t]{\textwidth}
        \centering
        \makebox[0.48\textwidth]{\textbf{1/3 Speed Factor}}
        \hfill
        \makebox[0.48\textwidth]{\textbf{1/2 Speed Factor}}

        \begin{minipage}[t]{0.48\textwidth}
            \centering
            \includegraphics[width=\textwidth]{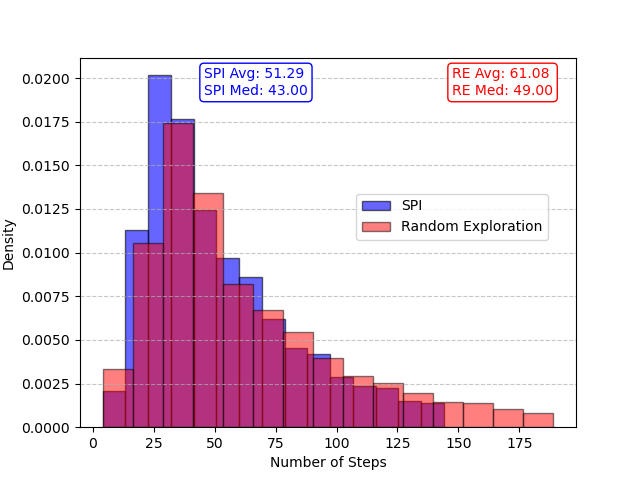}
        \end{minipage}
        \hfill
        \begin{minipage}[t]{0.48\textwidth}
            \centering
            \includegraphics[width=\textwidth]{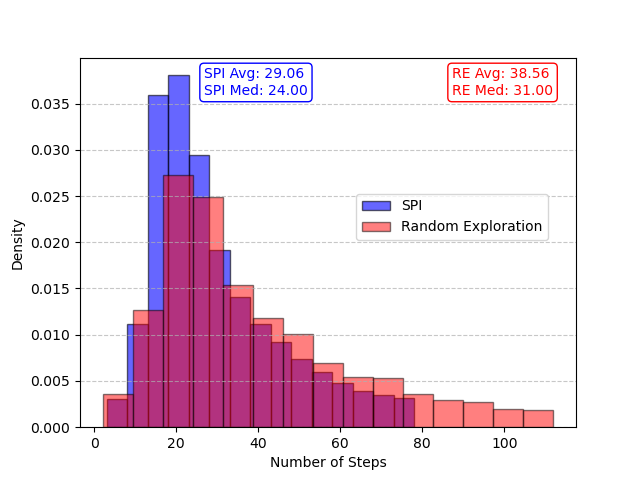}
        \end{minipage}

        \caption{\textbf{These figures illustrate and compare the failure steps taken during the first half of training at two different speed factors: 1/3 and 1/2.}}
        \label{fig:stepfailfirst}
    \end{minipage}
\end{figure}

\begin{figure}[H]
    \centering
    \begin{minipage}[t]{\textwidth}
        \centering
        \makebox[0.48\textwidth]{\textbf{1/3 Speed Factor}}
        \hfill
        \makebox[0.48\textwidth]{\textbf{1/2 Speed Factor}}

        \begin{minipage}[t]{0.48\textwidth}
            \centering
            \includegraphics[width=\textwidth]{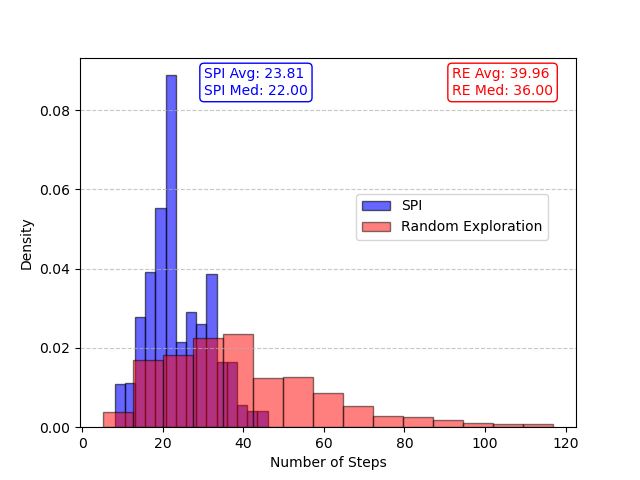}
        \end{minipage}
        \hfill
        \begin{minipage}[t]{0.48\textwidth}
            \centering
            \includegraphics[width=\textwidth]{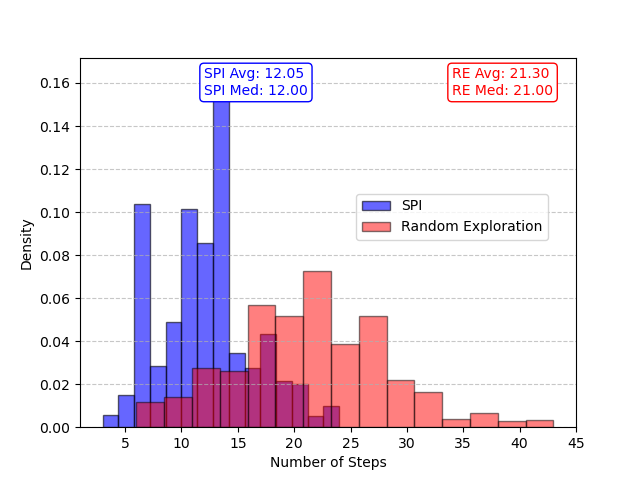}
        \end{minipage}

        \caption{\textbf{These figures illustrate and compare the failure steps taken during the second half of training at two different speed factors: 1/3 and 1/2.}}
        \label{fig:stepfailsecond}
    \end{minipage}
\end{figure}

For failure steps, SPI generally takes slightly fewer steps than random exploration (see Fig.~\ref{fig:stepfailall}). However, the comparison changes when examining failure steps at the beginning versus the end of training. Early on, the distribution of SPI and random exploration failure steps is similar to the overall failure step distribution (see Fig.~\ref{fig:stepfailfirst}). However, towards the end of training, when the success rate of each episode is high, the failure step distribution of SPI and random exploration diverges. When SPI fails during the second half of training, it consistently does so with a small step count, whereas random exploration failures during the second half of training are distributed over a wider range of steps and exhibit a higher average step count (see Fig.~\ref{fig:stepfailsecond}).

\subsubsection{Reward}

For both speed factors \(\frac{1}{2}\) and \(\frac{1}{3}\), agents receive similar rewards in the first few episodes. As training progresses, SPI consistently yields higher rewards at a speed factor of \(\frac{1}{3}\). A similar trend is observed for the \(\frac{1}{2}\) speed factor, though the difference in rewards between the two approaches is less pronounced.

\begin{figure}[htbp]
    \centering
    \begin{minipage}[t]{\textwidth}
        \centering
        \makebox[0.48\textwidth]{\textbf{1/3 Speed Factor}}
        \hfill
        \makebox[0.48\textwidth]{\textbf{1/2 Speed Factor}}
        
        \vskip\baselineskip 

        \begin{minipage}[t]{0.48\textwidth}
            \centering
            \includegraphics[width=\textwidth]{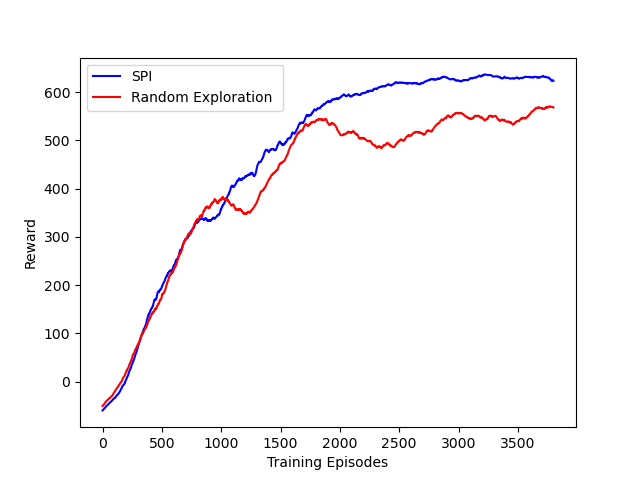}
        \end{minipage}
        \hfill
        \begin{minipage}[t]{0.48\textwidth}
            \centering
            \includegraphics[width=\textwidth]{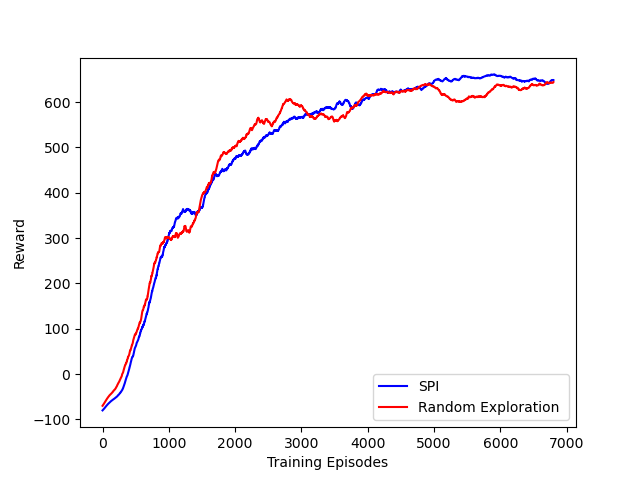}
        \end{minipage}
        
        \caption{\textbf{These figures illustrates and compares the reward obtained during training by agents using SPI or random exploration at two different speed factors: 1/3 and 1/2.}}
        \label{fig:reward}
    \end{minipage}
\end{figure}

\subsection{Discussion}

The results indicate that SPI can serve as an effective alternative to random exploration in agent-based self-organizing systems. Future research could explore SPI's adaptability to scenarios discouraging collaboration, as well as enhancements in SPI's data structure for more complex environments. For example, if there was a solution to find the optimal number of agents needed at any step, then we can update SPI to allow agents to adapt to different environments. But, this solution might take a lot more computational power and can be inaccurate. Future studies can expand on both the action and state spaces, as this study employs simplified versions that may limit the agents' access to information during training and execution. Broader action and state spaces could provide agents with a richer set of inputs and actions, potentially enhancing their adaptability and performance in complex environments.

SPI will eventually converge with random exploration given enough episodes. However, it is able to find a more efficient route faster than random exploration in the experiment. This is applicable for situations like swarm robotics where time efficiency and rapid learning is valuable. SPI is also scalable due to its decentralized nature. Agents follow a set global policy which allows for quick deployment for any number of agents. Furthermore, SPI requires no communication, implicit or explicit. This makes it valuable in systems with limited or no access to communication bandwidths during training.

The effectiveness of SPI depends on the environment. If many micro movements are needed to avoid obstacles -- when the most optimal action for agents is to not push the box --. then the proposed SPI will be ineffective. In that scenario, the inefficiency arises from the proposed SPI’s emphasis on maximizing box displacement every step. However, we have shown that the proposed SPI is very effective in an environment in which collaboration -- where each agent’s effort in pushing the box contributes positively - is essential for optimal success. Therefore, knowledge of the environment is critical to determine the structure and implementation of SPI.

\section{Conclusion and Future Work}

In this study, we tested the training effectiveness of agents in the box-pushing game. Our original RL technology SPI replaces the random exploration in MARL models to aid the coordination between agents in the box-pushing environment. This study presents SPI as a streamlined method for coordinating agent actions in multi-agent systems. After running computer simulations, we found that the effectiveness of the proposed SPI depends on the environment. In an environment that highly encourages all agents to apply force collaboratively to push the box, the proposed SPI trained faster and found a more efficient route than random exploration. By minimizing inter-agent conflicts, SPI facilitates efficient learning, enabling the system to achieve complex tasks more effectively.

In the future, we plan to test different SPI algorithms across randomized environments, defined only by a set of rules and underlying physics of the environment. This approach will allow us to evaluate the adaptability and robustness of SPI under diverse and unpredictable conditions.

\bibliographystyle{unsrt}
\bibliography{sample}

\end{document}